\documentclass{Interspeech2024}
\usepackage{multirow}
\newcommand{\specialcell}[2][l]{%
  \begin{tabular}[#1]{@{}l@{}}#2\end{tabular}}
\usepackage[acronym]{glossaries}
\usepackage{booktabs}
\newacronym{MAE}{MAE}{mean absolute error}
\newacronym{GT}{ground truth}{ground truth}
\usepackage{amsmath,amssymb}
\usepackage{mathtools}
\usepackage{relsize}
\usepackage{subfigure}
\usepackage{textcomp}
\usepackage{amsmath}
\usepackage{nicefrac}

\newcommand{\ddsp}{\mbox{DDSP-QbE }}
\usepackage[subtle]{savetrees}

\interspeechcameraready

\title{
Anonymising Elderly and Pathological Speech: Voice Conversion Using DDSP and Query-by-Example\vspace{-1em}}

\name[affiliation={1}]{Suhita}{Ghosh}
\name[affiliation={3}]{Melanie}{Jouaiti}
\name[affiliation={4}]{Arnab}{Das}
\name[affiliation={2}]{Yamini}{Sinha}
\name[affiliation={4}]{Tim}{Polzehl}
\name[affiliation={2}]{Ingo}{Siegert}
\name[affiliation={1}]{Sebastian}{Stober}

\address{
\normalsize
  $^1$Artificial Intelligence Lab, $^2$Mobile Dialog Systems, Otto-von-Guericke-University, Magdeburg, Germany \\
  $^3$School of Computer Science, University of Birmingham\\
  $^4$Speech and Language Technology, German Research Center for Artificial Intelligence (DFKI)
}
\email{\{suhita.ghosh,yamini.sinha,ingo.siegert,stober\}@ovgu.de, m.jouaiti@bham.ac.uk,
\{arnab.das,tim.polzehl\}@dfki.de}


\begin{document}

\maketitle
\begin{abstract}

Speech anonymisation aims to protect speaker identity by changing personal identifiers in speech while retaining linguistic content. Current methods fail to retain prosody and unique speech patterns found in elderly and pathological speech domains, which is essential for remote health monitoring. To address this gap, we propose a voice conversion-based method (DDSP-QbE) using differentiable digital signal processing and query-by-example. The proposed method, trained with novel losses, aids in disentangling linguistic, prosodic, and domain representations, enabling the model to adapt to uncommon speech patterns. Objective and subjective evaluations show that DDSP-QbE significantly outperforms the voice conversion state-of-the-art concerning intelligibility, prosody, and domain preservation across diverse datasets, pathologies, and speakers while maintaining quality and speaker anonymity. Experts validate domain preservation by analysing twelve clinically pertinent domain attributes.
\end{abstract}

\begin{figure*}[!ht] 
\centering
  \includegraphics[scale=0.88]{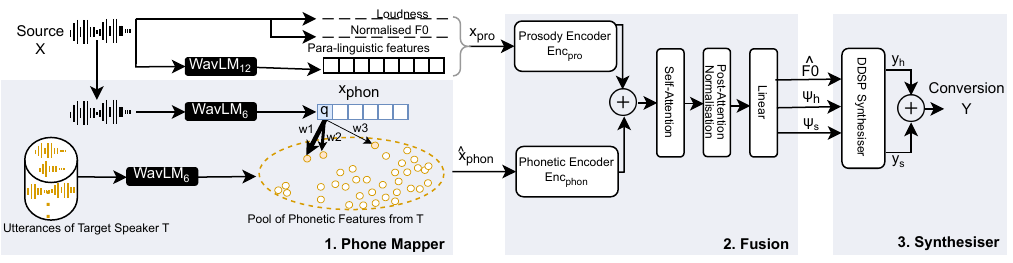}
   \vspace{-.75em}
  \caption{The proposed framework \ddsp. In the illustration, the number of matching candidates for computing $\hat{x}_{phon}$, is M=3.}
  \vspace{-0.5em}
  \label{fig:ddsp}
  \vspace{-0.75em}
\end{figure*}

\section{Introduction}
The widespread adoption of cloud-based speech technologies has made remote health monitoring more accessible for elderly individuals and those with speech disorders~\cite{kulkarni2022speech,ahmed2018early}.
However, since these speech recordings contain highly sensitive personal data, it becomes crucial to anonymise the speech before sharing the data across systems~\cite{schomakers2019privacy}.
Speech anonymisation aims to conceal the speaker's identity in recordings while maintaining linguistic content.
For elderly and pathological (non-standard) speech data, it is crucial to also preserve the prosody and unique speech patterns of the domain, such as the hoarseness in dementia patients~\cite{taylor2020age}, for further analysis, diagnosis, and tracking of age- or disease-related changes~\cite{liang2022evaluating}.

Voice conversion (VC)-based methods~\cite{sisman2020overview} have been successful in producing anonymised speech, where a source utterance is modified to sound like a \textit{target} speaker.
Most VC methods are based on generative adversarial network (GAN), trained with cycle-consistency loss, allowing training on non-parallel datasets~\cite{kaneko2018cyclegan}.
These methods generate a transformed spectrogram conditioned on the target speaker's embeddings, which are learnt jointly with the linguistic embeddings during training.
GAN-based methods overcome the buzzy voice problems caused by spectrum over-smoothing in variational autoencoder approaches~\cite{georges2022self}. This is attributed to the GAN's discriminator, which ensures that the generator produces realistic conversions matching the target speaker's style.
Conversely, techniques like KNN-VC~\cite{baas2023voice} opt for a strategy that eschews the direct learning of speaker or phonetic embeddings.
Instead, KNN-VC utilises self-supervised model-derived representations, where the features from the source are mapped to a target speaker using K-nearest neighbours.
Similarly, another method~\cite{nercessian2021end} used pre-trained speaker embeddings as one of the inputs in the differentiable digital signal processing (DDSP)~\cite{engel2020ddsp}-based framework.  
DDSP integrates traditional DSP elements, such as filters and synthesiser oscillators, into deep neural networks, with the neural network itself generating the parameters.
To the best of the authors' knowledge, our work is pioneering DDSP-based VC evaluation compared to other VC or anonymisation methods.
Most VC methods~\cite{walczyna2023overview} have primarily been evaluated on standard data, featuring speech from young and healthy adults, with an emphasis on the naturalness and intelligibility of conversions.
However, the recently introduced any-to-many method Emo-StarGAN~\cite{ghosh2023emo} extends focus to prosody, achieving emotion-preserving conversions by employing losses derived from both hand-crafted and deep learning (DL) generated para-linguistic features, as well as an adversarial emotion classifier.
However, the method fails to preserve the atypical speech patterns seen in stuttering~\cite{hintzanonymization}, a common form of non-standard data.

Thus, we propose `DDSP-QbE', an any-to-many VC method focused on preserving prosody and domain characteristics in speech anonymisation, even for unseen speakers from non-standard data.
Our method builds on recent advancements in query-by-example (QbE)~\cite{lea2023latent} and DDSP~\cite{wu2022ddsp}.
Our approach uses a subtractive harmonic oscillator-based DDSP synthesiser~\cite{wu2022ddsp}, inspired by the human speech production model~\cite{fant1981source}, to incorporate an inductive bias for effective learning with limited data. 
By leveraging QbE, we directly derive target phonetic representations from source speech, thus bypassing the need to learn these representations during training.
We introduce an inductive bias for prosody preservation by: (i) employing a novel loss function that utilises emotional speech to facilitate the separation of prosodic and linguistic features, and (ii) adding supplementary hand-crafted and DL-generated input features to the network, which have prosodic knowledge from the source utterance.
For domain preservation, we employ loss functions based on acoustic properties that are crucial in clinical evaluations of voice disorders~\cite{nishikawa2022analysis}.
In addition to an objective evaluation, we conduct an in-depth subjective assessment of domain preservation, with speech pathologists assessing the retention of twelve clinically recognised measures for voice disorder.
This analysis offers key insights, highlighting which domain aspects are preserved and which are not.
Both objective and subjective analyses demonstrate DDSP-QbE's ability to anonymise speech while retaining prosodic and clinically pertinent domain features.

\section{Differentiable Digital Signal Processing}
In the DDSP framework, a synthesiser generates speech, with its parameters predicted by a neural network, enabling end-to-end training.
However, this necessitates that the synthesiser's components are differentiable to enable back-propagation.
A subtractive-based synthesiser model~\cite{wu2022ddsp} is a harmonic-plus-noise model, which decomposes a monophonic sound into two components, harmonic~$y_h$ and stochastic~$y_s$, through two stages.
In the first stage, the synthesiser approximates ${y}_h$ by a fundamental frequency (F0)-constrained sawtooth signal and derives the unfiltered harmonic component as $\Tilde{y}_h(t) = \sum^J_{j=1}\frac{1}{j}sin(\phi_j(t))$, where $\phi_j(t)$ is the phase for the $j^{th}$ harmonic at the $t^{th}$ time instant.
On the other hand, the unfiltered stochastic component $\Tilde{y}_s$ is derived from a uniform noise signal $\eta \in [-1,1]$.
In the second stage, ${y}_h$ and ${y}_s$ are acquired through individual spectral filtering using linear time-varying finite impulse response (LTV-FIR) filters $\psi_h \in \mathbb{R}^{F_h}$ and $\psi_s \in \mathbb{R}^{F_s}$, respectively~\cite{wu2022ddsp}.
The final audio is obtained as $Y =  y_h + y_s$.
    

\section{\ddsp}
The state-of-the-art VC approach Emo-StarGAN~\cite{ghosh2023emo} successfully retains prosody during anonymisation for standard data. However, it struggles to preserve the distinctive speech patterns, or domain features characteristic of elderly and pathological speech, such as roughness, strain, or breathiness~\cite{taylor2020age}.
Our method aims to overcome these challenges by isolating these unique domain-specific patterns from personal speaker traits, even in situations with limited data.

\subsection{Framework}
\label{sec:fw}
As portrayed in Fig.~\ref{fig:ddsp}, the proposed framework comprises three components: Phone Mapper, Fusion, and Synthesiser, with only the latter two requiring training.

\textbf{Phone Mapper:} Given a source utterance $X$, we obtain the phonetic representations $\hat{x}_{phon}$ for a \textit{target} speaker from Phone Mapper.
These representations maintain the linguistic content of the source utterance while sounding like the target speaker.
Previous works~\cite{baas2023voice,chen2022wavlm} have shown that representations from the 6\textsuperscript{th}~layer of the self-supervised model WavLM-Large~(WLM\textsubscript{6}) serve as a promising candidate to derive \textit{latent} phonetic features as they achieved high performance in phone discrimination tasks~\cite{dunbar2022self}.
Furthermore, similar-sounding phones tend to be closer together in this latent phone space~\cite{baas2023voice}.
With this understanding, we derive the target-phone features from the source-phone features through a QbE scheme, similar to the previous works~\cite{baas2023voice,lea2023latent}.

Initially, a pool of phone representations is generated per target speaker using WLM\textsubscript{6}, computed frame-wise for all available utterances.
For a source utterance $X$, phonetic representations $x_{phon}$ are derived in a similar manner using WLM\textsubscript{6}.
Each source-phone representation acts as a query $q \in x_{phon}$ (shown in Fig.~\ref{fig:ddsp}), which is replaced with $\hat{q} \in \hat{x}_{phon}$.
$\hat{q}$ is computed as the weighted average of top-$M$ phone representations similar to $q$ from the target speaker's pool.
Specifically, it is calculated as $\hat{q} = \frac{\sum_{i=1}^M m_i*w_i}{\sum_{i=1}^M w_i}$, where $m_i$ represents the selected phone representation, and $w_i$ is its corresponding weight.
The weights $\{w_i\}_{i=1}^M$ are determined by applying a softmax function to the inverse of the cosine distance between $q$ and the $\{m_i\}_{i=1}^M$ candidates.
The weighting scheme is utilised to reduce the impact of outliers, thereby ensuring that phonetic representations closer in distance to $q$ are given more importance.

\textbf{Fusion:}
This component generates the parameters needed by the DDSP synthesiser, as shown in Fig.~\ref{fig:ddsp}.
It has been found that the phonetic representations from WLM\textsubscript{6} are intertwined with para-linguistic or prosodic cues~\cite{baas2023voice}.
This implies that during the \textit{mapping phase}, the prosodic information is also replaced along with phonetic features, which is undesirable as we aim to preserve the prosody from the source.
To address this `prosody leakage' issue, 
we provide the network with prosodic features $x_{pro}$, extracted directly from the source utterance, along with the mapped phonetic features $\hat{x}_{phon}$ from the Phone~Mapper, as shown in Fig.~\ref{fig:ddsp}.
The prosodic features $x_{pro}$ are a combination of hand-crafted and DL-generated features, which are correlated with prosodic cues. 
The hand-crafted features considered are loudness and sample-wise z-normalised logarithmic F0 contour, aimed at capturing the speaker-independent pitch variations.
Recent analyses \cite{zhu2023deep,li2023exploration} have shown the middle (12\textsuperscript{th}) layer of WavLM-Large (WLM\textsubscript{12}) to perform well for para-linguistic related tasks, such as emotion classification.
Therefore, we consider the representations from WLM\textsubscript{12} as the prosody-correlated deep feature as prosody provides important cues to emotion~\cite{cao2014prosodic}.

Initially, $\hat{x}_{phon}$ and $x_{pro}$ are fed to their individual branches, phonetic encoder Enc\textsubscript{phon} and prosody encoder Enc\textsubscript{pro} respectively, as shown in Fig.~\ref{fig:ddsp}.
Each branch comprises two 1D convolutions with ReLU activation followed by group normalisation.
The outputs from both branches are combined through element-wise addition and passed through a stack of three self-attention layers.
Subsequently, this is followed by a shallow convolution stack with post-attention normalisation.
Finally, a linear layer is used with dimensions matching the number of parameters $\Theta$, required by the synthesiser.
The architecture of the Fusion component is akin to the small Conformer architecture \cite{gulati2020conformer}, which has demonstrated efficacy in capturing both local and global contexts in a sequence of acoustic features.

\textbf{Synthesiser}:
We integrate the subtractive synthesiser proposed in~\cite{wu2022ddsp} into our framework. The synthesiser produces the conversion $Y$ using the parameters $\Theta$ derived from the Fusion module, where $\Theta=\{\widehat{F0}, \psi_h, \psi_s\}$.
In our work, we incorporate the network predicted F0 ($\widehat{F0}$) unlike in the original DDSP work~\cite{engel2020ddsp}, which incorporates an additional inductive bias in the network and drives it to produce F0-consistent speech.
\vspace{-0.5em}
\subsection{Domain and Prosody-Aware Losses }
Relying solely on multi-resolution spectral losses, as employed in previous works~\cite{nercessian2021end,engel2020ddsp,wu2022ddsp}, ensures high fidelity in the reconstruction, but fails to guarantee the preservation of non-linguistic features.
To address this, we incorporate additional losses during training.

\textbf{Jitter and Shimmer} are clinically acclaimed indicators for assessing voice disorders~\cite{sripriya2017non}.
Jitter refers to the variation or irregularity in the timing of consecutive periods of F0, reflecting the instability in vocal fold vibrations~\cite{teixeira2016algorithm}.
Therefore, jitter can be used to assess `shakiness' or `unsteadiness' correlates in the voice.
We consider the jitter of the five-point period perturbation quotient ($j_{ppq5}$) as it is widely used in clinical studies for its ability to provide a more consistent assessment by considering neighbouring periods~\cite{barsties2023advances}.
$T_i$s are the extracted F0 period lengths and $N$ is the number of F0 periods.
We calculate jitter loss $L_{jit}$ as the mean absolute error (MAE) between $j_{ppq5}$ computed from the source $X$ and the conversion $Y$.
\begin{equation}
j_{ppq5} = \frac{\frac{1}{N-1}\sum_{i=2}^{N-2}|T_i - \frac{1}{5}\sum_{n=i-2}^{i+2}T_n|}{\frac{1}{N}\sum_{i=1}^NT_i} * 100%
\label{eq:ppq5}
\end{equation}

Shimmer, on the other hand, captures amplitude irregularities, which can be used to capture `roughness' and `breathiness' correlates in the voice~\cite{barsties2023advances,barcelos2018multidimensional}.
We consider the \textit{local} shimmer $s_{loc}$, which is used in the clinically recognised measure to assess breathiness, acoustic breathiness index (ABI)~\cite{latoszek2017acoustic}.
$s_{loc}$ is computed by taking the absolute average difference between the amplitudes of consecutive periods, and then dividing it by the average amplitude~\cite{barsties2023advances}.
We calculate the shimmer loss $L_{shim}$ as the MAE between the $s_{loc}$ extracted from the source $X$ and the converted speech $Y$ samples.

\textbf{Prosody Leakage}
from the mapping phase is addressed by introducing a loss formulation using emotional utterances.
The Enc\textsubscript{phon} encoder is specifically designed not to capture prosodic representations but only phonetic representations that sound like the target speaker.
Thus, Enc\textsubscript{phon} should generate comparable representations for two utterances ($X_1$ and $X_2$) by the same speaker, containing the same linguistic content but delivered with different emotions.
To quantify the disparity between these representations, we introduce the loss $L_{pro}$, as depicted in Equation~\ref{eqn:emo}.
This loss guides the encoder Enc\textsubscript{phon} to capture non-prosodic representations, thereby facilitating the disentanglement of prosodic and non-prosodic features.
\begin{equation}
\hspace{-0.2em}
\label{eqn:emo}
    L_{pro} = \left|\operatorname{Enc_{phon}}(\operatorname{WLM_6}(X_{1})) - \operatorname{Enc_{phon}}(\operatorname{WLM_6}(X_{2}))\right|
\end{equation}

\noindent\textbf{Training Objectives}: We train the \ddsp model with the multi-resolution spectral loss $L_{s}$ and F0-related loss $L_{f0}$, as done in \cite{wu2022ddsp}, along with our proposed losses.
Therefore, we train the model \ddsp using the objective function shown in Equation~\ref{eqn:loss}, where $\lambda_{s}, \lambda_{jit}, \lambda_{shim}, \lambda_{pro}$ and $\lambda_{f0}$ are hyper-parameters.
\begin{equation}
    \hspace{-0.6em}
\label{eqn:loss}
    Loss = \lambda_{s}L_{s} + {\lambda_{jit}L_{jit} + \lambda_{shim}L_{shim}} + \lambda_{pro}L_{pro} + \lambda_{f0}L_{f0}
\end{equation}

\section{Experiment and Results}
\begin{table*}[htbp]
\caption{Objective evaluation results with 95\% confidence intervals are presented. `Domain Pr.' indicates Domain Preservation. The `Type' column specifies special cases, such as the source speaker's domain, source and target gender groups, or `All', which includes all sub-groups.}
\resizebox{\textwidth}{!}
{
 \begin{tabular}{l|l|cc|ll|ll|ll|ll}
 \toprule
 {\textbf{\specialcell{Source}}} & {\textbf{\specialcell{Type}}} &
 \multicolumn{2}{c|}{\textbf{Domain Pr.} [\%] $\uparrow$}  & \multicolumn{2}{c|}{\textbf{PCC} [$\times 10^{2}$] $\uparrow$} & \multicolumn{2}{c|}{\textbf{pMOS} $\uparrow$} & \multicolumn{2}{c|}{\textbf{CER} [\%] $\downarrow$} & \multicolumn{2}{c}{\textbf{EER} [\%] $\uparrow$} \\ 
\cline{3-12}

{} & {} & {\textbf{Emo}} & {\textbf{DDSP-QbE}} & {\textbf{Emo}} & {\textbf{DDSP-QbE}}& {\textbf{Emo}} & {\textbf{DDSP-QbE}}& {\textbf{Emo}} & {\textbf{DDSP-QbE}}& {\textbf{Emo}} & {\textbf{DDSP-QbE}}\\ 
\midrule
\multirow{3}{*}{\textbf{All conversions}} & {All} & 67.8$\pm$1.2& \textbf{78.1}$\pm$2.3& {68.1$\pm$0.7} &{\textbf{77.4}$\pm$0.6} &{2.41$\pm$0.02} & {\textbf{3.39}$\pm$0.02}&{18.15$\pm$1.00} &{\textbf{1.04}$\pm$0.09}  &{\textbf{50.16}$\pm$0.03} &48.98$\pm$0.07\\
 \cmidrule{2-12}
  & {Different gender}& 64.7$\pm$0.9&\textbf{76.2}$\pm$2.5  &{67.6$\pm$1.0}  &{\textbf{75.4}$\pm$0.9} &{2.41$\pm$0.03} &{\textbf{3.36}$\pm$0.03} &{22.57$\pm$1.41} &{\textbf{1.05}$\pm$0.12} &- &-
  \\
& {Same gender}  &70.9$\pm$1.4 &\textbf{80.0}$\pm$2.1 &{68.6$\pm$0.9}  &{\textbf{79.4}$\pm$0.8}  &{2.42$\pm$0.03} &{\textbf{3.41}$\pm$0.02}  &{13.73$\pm$1.43} &{\textbf{1.02}$\pm$0.12} &- &-\\

 \midrule
 \multirow{3}{*}{\textbf{Elderly $\rightarrow$ SD}}  & {All} &61.3$\pm$0.7 &\textbf{79.7}$\pm$1.7 & {55.1$\pm$1.2}  &{\textbf{70.6}$\pm$1.4} &{2.28$\pm$0.04} &{\textbf{3.24}$\pm$0.03} &{24.8$\pm$1.61} &{\textbf{2.41}$\pm$0.21}  &40.62$\pm$0.11 &{\textbf{43.41}}$\pm$0.08
\\
 \cmidrule{2-12}
 & {Dementia} &63.3$\pm$0.9 &\textbf{78.9}$\pm$2.5 & {53.5$\pm$1.7}  &{\textbf{68.4}$\pm$2.0} &{2.31$\pm$0.05} &{\textbf{3.29}$\pm$0.04} &{23.36$\pm$2.26} &{\textbf{2.40}$\pm$0.29} &{38.24$\pm$0.08} &{\textbf{40.41}$\pm$0.05} \\ 
 
 & {Healthy}&59.3$\pm$0.8 &\textbf{80.5}$\pm$0.9 &{56.6$\pm$1.7}  &{\textbf{72.8}$\pm$1.8}  &{2.24$\pm$0.05} &{\textbf{3.20}$\pm$0.05}  &{26.24$\pm$2.29} &{\textbf{2.42}$\pm$0.31} &{43.40$\pm$0.06} & {\textbf{44.39}$\pm$0.08}
\\
 \midrule
 \multirow{6}{*}{\textbf{Stuttering $\rightarrow$ SD}} & {All} &64.3$\pm$2.7 &\textbf{77.8}$\pm$2.2 &{{80.0$\pm$0.7}}  &{\textbf{84.2}$\pm$0.5} &{2.41$\pm$0.04} &{\textbf{3.33}$\pm$0.03} &{19.07$\pm$4.01} &{\textbf{0.13}$\pm$0.05} &{\textbf{48.99}$\pm$0.02} &{48.85$\pm$0.05}
 \\

\cmidrule{2-12}
 
 & {Block} &59.7$\pm$3.7 &\textbf{78.3}$\pm$2.2  &{80.3$\pm$1.7}  &{\textbf{83.8}$\pm$1.0} &{2.41$\pm$0.08} &{\textbf{3.37}$\pm$0.07} &{16.55$\pm$4.00} &{\textbf{0.08}$\pm$0.08}&- &-
\\
 
 & {Word Repetition} &73.2$\pm$1.7 &\textbf{76.3}$\pm$2.1 &{78.9$\pm$1.6}  &{\textbf{83.6}$\pm$0.9} &{2.38$\pm$0.09} &{\textbf{3.34}$\pm$0.06} &{16.91$\pm$3.93} &{\textbf{0.15}$\pm$0.10}&- &-
\\
 
 & {Sound Repetition} &49.8$\pm$4.7 &\textbf{78.0}$\pm$2.2 &{80.3$\pm$1.6}  &{\textbf{84.6}$\pm$1.0} &{2.37$\pm$0.10} &{\textbf{3.22}$\pm$0.07}  &{22.50$\pm$3.94} &{\textbf{0.20}$\pm$0.12}&- &-
\\
 
 & {Interjection} &74.5$\pm$0.2 & \textbf{78.2}$\pm$2.3 &{79.1$\pm$1.7}  &{\textbf{83.6}$\pm$1.1}  &{2.47$\pm$0.08} &{\textbf{3.37}$\pm$0.07} &{20.32$\pm$3.99} &{\textbf{0.08}$\pm$0.08}&- &-
\\
\midrule
\multirow{1}{*}{\textbf{SD $\rightarrow$ SD}}  & {All}  &\textbf{77.8}$\pm$1.6 &76.8$\pm$1.7  &{79.9$\pm$1.0}  &{\textbf{81.4}$\pm$0.4} &{3.55$\pm$0.04} &{\textbf{3.88}$\pm$0.02} &{10.58$\pm$0.85} &{\textbf{0.57}$\pm$0.09}  & {\textbf{52.29}$\pm$0.05}&{51.66$\pm$0.04}
\\
 \bottomrule
\end{tabular}
}

\label{tab:objective}
\end{table*}

\begin{figure*}[!ht] 
\vspace{-0.25em}
  \includegraphics[width=\textwidth]{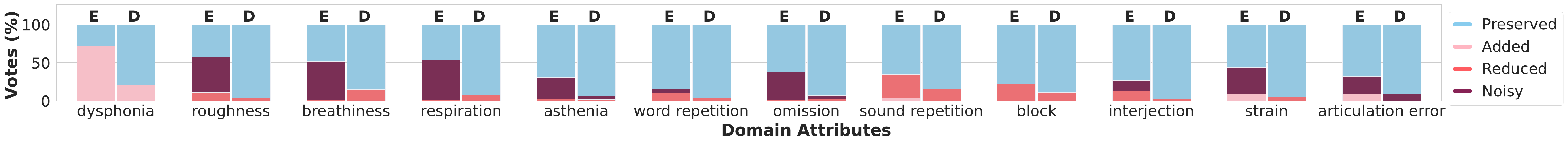}
 \vspace{-2em}
  \caption{Subjective evaluation results for domain preservation. Illustrates the percentage of each domain attribute being `preserved', `added' (or increased), `removed' (or reduced), or `noisy' (could not be assessed), for each model. `E' represents Emo, and `D' stands for DDSP-QbE.}
  \vspace{-1.5em} 
  \label{fig:subjective}
\end{figure*}
Due to the dearth of anonymisation methods for non-standard data, we use Emo-StarGAN (denoted as `Emo'), known for emotion preservation as our baseline, which has been evaluated on German stuttering speech~\cite{hintzanonymization}.

\textbf{Three English datasets} are
considered for training and evaluation: (i) ESD~\cite{zhou2022emotional}: standard data having annotations for 5 emotion classes, (ii) ADReSS\textsubscript{0}~\cite{luz2021detecting}: speech from healthy and dementia-diagnosed elderly speakers, and (iii) Sep-28k~\cite{lea:2021}: stuttering speech, containing annotations for 4 types of stuttering: block, interjection, word and sound repetitions.
All utterances are resampled to 16 kHz and randomly split into training, validation, and test sets in proportions 0.8/0.1/0.1, respectively, ensuring no speakers overlap across the sets.
We train and evaluate all the models on the same splits for a fair comparison.
For ESD, we consider utterances from 6 speakers for training.
For the non-standard datasets, we consider 20 speakers each from ADReSS\textsubscript{0} and Sep-28k, totalling $\approx$2.5 hours of data.
Only excerpts featuring elderly and pathological speech are retained, guided by the dataset annotations.
Speaker selections across all datasets are randomised, ensuring an even distribution of healthy, non-healthy, various pathologies and genders.
Each model is trained using Adam optimiser with a learning rate of 0.002 for 150 epochs and batch size of 128, taking $\approx$36 hours to complete on an Nvidia A100 \SI{80}{\giga\byte} GPU.
WavLM representations for training are produced from 2-second audio.
We use $J$=150 harmonics and the filter lengths as $F_h$ = 176 and $F_s$ = 80, and 5 resolutions $R=\{2^i\}_{i=6}^{10}$ for spectral loss $L_{s}$, with 75\% overlapping among neighbouring frames.
We set $\lambda_{ms}$=$\lambda_{f0}$=1.0, $\lambda_{jit}$=10, $\lambda_{shim}=0.1$ and $\lambda_{pro}$=0.1.
\ddsp generate conversions faster than real-time, considering $M$=4 candidates from a phonetic feature pool, which is created for each target speaker from around 5 minutes of their utterances.
We train a HiFiGAN vocoder on the training split, as described in~\cite{ghosh2023emo}, which is utilised by Emo to produce conversions.
The remaining intricate training details and demo audio samples can be found online\footnote{https://github.com/suhitaghosh10/ddsp-qbe.git}.\\
\textbf{Evaluation Setup}:
We perform both objective and subjective evaluations for 4 source~$\rightarrow$~target scenarios: (i) Elderly+Healthy~$\rightarrow$~SD, (ii) Elderly+Dementia~$\rightarrow$~SD, (iii) Stuttering~$\rightarrow$~SD, and (iv) SD~$\rightarrow$~SD, where SD denotes standard data.
In all scenarios, the target speaker is chosen from SD, mirroring real-life situations due to the widespread availability of standard data.
For evaluation, we use source utterances from the test split and randomly select 1000 conversions for each scenario, ensuring a balanced distribution across genders and types of pathologies.

\textbf{Objective Evaluation}: We assess domain preservation by classifying stuttering types and dementia. The classifiers are trained only on the original utterances in the training split, as done in~\cite{ghosh2023emo}.
We compare the class predicted for the converted speech with that of the original speech, considering the latter as the ground truth.
Specifically for SD$\rightarrow$SD, we analyse whether pathologies such as dementia or stuttering are inadvertently introduced during the conversion process.
For other performance metrics, we follow the methods detailed in previous work~\cite{ghosh2023emo}: assessing overall quality through the predicted mean opinion score (pMOS)~\cite{andreev2022hifi++}, prosody preservation via the pitch-correlation coefficient (PCC), intelligibility by measuring the character error rate (CER) using transcriptions from the Whisper medium-English model~\cite{radford2023robust}, and the strength of anonymisation through the equal error rate (EER).

\textbf{Subjective Evaluation}:
We embrace two kinds of user studies\footnote{conducted on Crowdee: \url{https://www.crowdee.com}}, considering 160 randomly selected conversions per model due to the extensive time and cost involved in evaluating all of them.
Each audio clip lasts for 4-13 seconds.
The raters for the studies were unaware whether the samples were original or synthetically generated.

In the first study, two speech-language pathologists assessed the preservation of 12 domain attributes.
These measures are typically used clinically to detect speech disorders~\cite{martinez2021ten,perez2016stuttering}: (i) Dysphonia: following the GRBAS scale~\cite{hirano1981clinical}, which provides an assessment of the severity level of a speech disorder, (ii) roughness: measures raspiness or harshness in voice, (iii) breathiness: measures lack of clarity in phonation, (iv) abnormal respiration, (v) articulation error, (vi) word repetition such as `I will [will] go', (vii) sound repetition such as `I am [pr-pr-pr-]prepared', (viii) omission or made-up words, (ix) block: unnatural pause or gasps of air, (x) interjection or filler-words such as `um' or `uh', (xi) strain: excessive effort or tension in phonation, and (xii) asthenia: lack of strength in the voice.

In the second study, 87 English-speaking participants assessed prosody, naturalness and anonymisation, as done in~\cite{ghosh2023emo}.
For prosody preservation, subjects compared the rhythm and intonation of a source utterance to the conversions by Emo and DDSP-QbE (ABX test), disregarding quality and content, and choosing the more similar or `both equal' option.
They rated naturalness on a 5-point MOS scale from `bad' to 'excellent'.
For anonymisation, raters marked speaker similarity on a 5-point scale (1: different, 5: similar), after listening to a converted sample and another utterance from the source speaker.
Each test was assessed by at least 3 raters, with those failing hidden traps twice excluded from analysis.\\
\textbf{Results~and~Discussion}:
Table~\ref{tab:objective} indicates that DDSP-QbE significantly surpasses Emo concerning all metrics except for anonymisation, where the EER scores are comparable.
However, in subjective assessment (refer to Table~\ref{tab:subj}), \ddsp conversions were perceived as less similar to the source speaker compared to those from Emo.
Concerning intelligibility, \ddsp outperforms Emo significantly, as indicated by the CER scores (p $<$ 0.001, paired t-test).
\begin{table}[h]
\caption{Subjective MOS and speaker similarity values with 95\% confidence intervals. The Prosody column indicates the percentage of votes each model received. Source utterances' mean MOS is 3.4.}

\resizebox{0.47\textwidth}{!}
{
 \begin{tabular}{l|ll|cc|ll}
 \toprule
\multirow{2}{*}{\textbf{Type}} &\multicolumn{2}{c|}{\textbf{MOS} $\uparrow$} & \multicolumn{2}{c|}{\textbf{Prosody [\%]} $\uparrow$} & \multicolumn{2}{c}{\textbf{Speaker Similarity} $\downarrow$} \\ 
\cline{2-7}
 {} &{\textbf{Emo}} & {\textbf{DDSP-QbE}}&{\textbf{Emo}} & {\textbf{DDSP-QbE}} & {\textbf{Emo}} & {\textbf{DDSP-QbE}} \\ 
\midrule
All &3.54$\pm$0.17 &\textbf{3.56}$\pm$0.13 &34.8 &\textbf{65.2} &{3.01$\pm$0.38} &{\textbf{2.51}$\pm$0.41}  \\
\midrule
{Different gender} & 3.49$\pm$0.18&\textbf{3.55}$\pm$0.25 &25.7  & \textbf{74.3}&{2.70$\pm$0.54}& {\textbf{2.21}$\pm$0.57} \\
{Same gender}  & 3.53$\pm$0.24& \textbf{3.64}$\pm$0.18&43.9 &\textbf{56.1} &{3.22$\pm$0.60} & {\textbf{2.98}$\pm$0.35} \\
 \bottomrule
\end{tabular}
}
\label{tab:subj}
\end{table}
Emo's reduced intelligibility stems from its challenge in adapting to uncommon speech patterns, such as irregular fricatives or plosives leading to prolongations, and repetitions of sounds like `[pr-pr-pr-]prepared', resulting in less clear distinctions between vowels and consonants.
This issue also adversely affects Emo's scores for prosody preservation, as seen in subjective results in Table~\ref{tab:subj}.
Further, Table~\ref{tab:objective} shows that both models struggle more with maintaining prosody in elderly speech than in stuttering speech, likely due to increased jitter and vocal tremors in elderly speech~\cite{schultz2023cross}. Nonetheless, DDSP outperforms Emo, achieving a higher mean PCC of 70.6 versus Emo's 55.1 for elderly speech.
Emo is more inclined than \ddsp to substitute the domain traits, such as roughness, breathiness, respiration, sound repetition, and blocks, with noise, as seen in Fig.~\ref{fig:subjective}.
This lowers the quality of Emo's conversions, leading to lower MOS scores compared to DDSP-QbE.
Most of the Emo's conversions were perceived to have an increased severity of voice disorder compared to the original utterance, affecting the Dysphonia score, as seen in Fig.~\ref{fig:subjective}.
\ddsp successfully preserves most domain attributes but faces difficulties with breathiness, block and sound repetition, as seen in Fig.~\ref{fig:subjective}. Specifically, \ddsp occasionally fails to accurately simulate the airflow release interruption in plosive sounds /p/, /b/, /t/, /d/, typical of stuttering speech.
This suggests a need for advanced modelling techniques, incorporating metrics like cepstral peak prominence (CPP) and Acoustic Breathiness Index (ABI)~\cite{latoszek2017acoustic}.

The ablation study presented in Table~\ref{tab:ablation} shows that removing $L_{pro}$ diminishes prosody preservation, while the individual removal of jitter and shimmer losses compromises the model's domain preservation capabilities.
Interestingly, the absence of each component reduces the intelligibility of the conversions as well.

\begin{table}[!h]
\caption{Ablation results with 95\% confidence intervals shown.}
\resizebox{0.47\textwidth}{!}
{
 \begin{tabular}{lcllll}
 \toprule
     {\textbf{\textbf{Method}}} &
     {\textbf{Domain Pr. [\%] $\uparrow$}} &{\textbf{PCC [$\times 10^{2}$] $\uparrow$}} & {\textbf{pMOS} $\uparrow$} & {\textbf{CER} [\%] $\downarrow$} & {\textbf{EER} [\%] $\uparrow$}   \\
     \midrule
    {Full DDSP-QbE}& {78.1$\pm$2.3} &{77.4$\pm$0.6} &{3.39$\pm$0.02} &{1.04$\pm$0.09} & 48.98$\pm$0.07\\
    \midrule
    {$\lambda_{pro}=0$ }& {73.7$\pm$1.3}&{67.9$\pm$0.5} &{3.40$\pm$0.03} &{4.56$\pm$0.53}& 50.99$\pm$0.09\\
    {$\lambda_{jit}=0$ }&{69.7$\pm$0.9}& {74.1$\pm$0.7} &{3.36$\pm$0.02} &{6.81$\pm$0.51}& 49.88$\pm$0.05
\\
    {$\lambda_{shim}=0$ }&{70.7$\pm$1.1}& {73.6$\pm$0.6} &{3.35$\pm$0.02} &{7.89$\pm$0.54}& 50.60$\pm$0.03 \\
     \bottomrule
\end{tabular}
}
\vspace{-1.5em}
\label{tab:ablation}
\end{table}

\section{Conclusion}
We propose the first speech anonymisation technique that successfully maintains the prosody and distinct speech characteristics prevalent in the elderly and pathological speech, while anonymisation.
Our approach utilises a subtractive DDSP synthesiser combined with query-by-example (QbE), possesses only \textbf{$0.4\%$} of the trainable parameters of Emo, and is trained on just $\approx$2.5 hours of data.
Despite this, it shows a superior ability to generalise to rare speech patterns, showing the effectiveness of the proposed inductive biases.
The detailed subjective assessments, including the one focusing on clinically relevant attributes, indicate that DDSP-QbE substantially surpasses the baseline in preserving both prosody and domain-specific traits across diverse speech patterns seen in non-standard data, target speakers, and genders.
Looking ahead, our goal is to improve the retention of complex features such as breathiness and to adapt our approach to additional languages and speech disorders.
\section{Acknowledgements}
This research has been supported by the Federal Ministry of Education and Research of Germany through project Emonymous (project number S21060A) and Medinym (focused on AI-based anonymisation of personal patient data in clinical text and voice datasets).

\bibliographystyle{IEEEtran}
\bibliography{ddsp_v2}

\end{document}